\documentclass{article}

\usepackage[preprint]{neurips_2022}

\usepackage[utf8]{inputenc} 
\usepackage[T1]{fontenc}    
\usepackage{hyperref}       
\usepackage{url}            
\usepackage{booktabs}       
\usepackage{amsfonts}       
\usepackage{nicefrac}       
\usepackage{microtype}      
\usepackage{xcolor}         
\usepackage{graphicx}

\usepackage{caption}

\usepackage{algorithm}
\usepackage{algorithmic}
\usepackage{graphicx}
\usepackage{subcaption}

\usepackage{xspace}
\usepackage{amsmath,amssymb,mathtools,amsthm}
\usepackage{multirow}
\usepackage{pifont}
\usepackage{xparse}
\usepackage{pythonhighlight}
\usepackage{array}
\newcolumntype{H}{>{\setbox0=\hbox\bgroup}c<{\egroup}@{}}

\title{Investigating Prompt Engineering in Diffusion Models}

\author{
  Sam Witteveen \\
  Red Dragon AI \\
  {\tt sam@reddragon.ai} \\
\And
  Martin Andrews \\
  Red Dragon AI \\
  {\tt martin@reddragon.ai} \\
}

\begin{document}

\maketitle

\begin{abstract}
With the spread of the use of Text2Img diffusion models such as DALL-E 2, Imagen, Mid Journey and Stable Diffusion, one challenge that artists face is selecting the right prompts to achieve the desired artistic output.
We present techniques for measuring the effect that specific words and phrases in prompts have,
and (in the Appendix) present guidance on the selection of prompts
to produce desired effects%
.
\end{abstract}

\section{Introduction \& Related Works}

The use of diffusion models for text guided image generation has 
become an important paradigm
in the last year with the introduction of models such as GLIDE (\cite{Nichol2022GLIDETP}), DALL-E 2 (\cite{DALLE2}), Imagen (\cite{Imagen}) and Latent Diffusion (\cite{LatentDiffusion}). 




In early work, Diffusion models were often conditioned on dataset such as Imagenet (\cite{deng2009imagenet}).
Recent models \cite{DALLE2,Imagen} have moved away from classifier guidance (\cite{DDPM}) 
to guiding their output via the embeddings resulting from text input to 
language models such as CLIP (\cite{CLIP}). 

The Stable Diffusion model (released under a permissive license in August 2022) is a Latent Diffusion Model (\cite{LatentDiffusion}) trained on 512x512 images from the LAION Aesthetics dataset of 2 Billion images (a subset of the LAION-5B database \cite{schuhmann2022laionb}). 

\section{Methodology}



In the following, we divide the textual prompt into its two main components :
(a) The physical and factual content of the image (eg. ``A Mainecoon cat kneeling''); 
and (b) the stylistic considerations in the way the physical content is displayed 
(eg. lighting, and other descriptors of style of the image).



Our experiments show that it is possible to measure which words and phrases 
are likely to fall into each of the categories and then 
measure the effect of these words and phrases on the generated image.


\begin{table}[!t]
\begin{minipage}[b]{0.5\linewidth}
\centering
\begin{tabular}{l r r r r}
    \hline \\[0.1ex]
    Prompt Collection & LPIPS & VGG & CLIP \\[1.2ex]
    \hline \\
    Descriptors & 0.664 & 0.646 & 0.839 \\[1ex]
    Nouns & 0.512 & 0.513  & 0.200 \\[1ex]
    Artists & 0.465 & 0.530  & 0.627 \\[1ex]
    \hline \\[0.1ex]
    \end{tabular}
    \caption{Image and Text average similarity}
    \label{table:image-similarity}
\end{minipage}\hfill
\begin{minipage}[b]{0.5\linewidth}
\centering
\includegraphics[width=55mm]{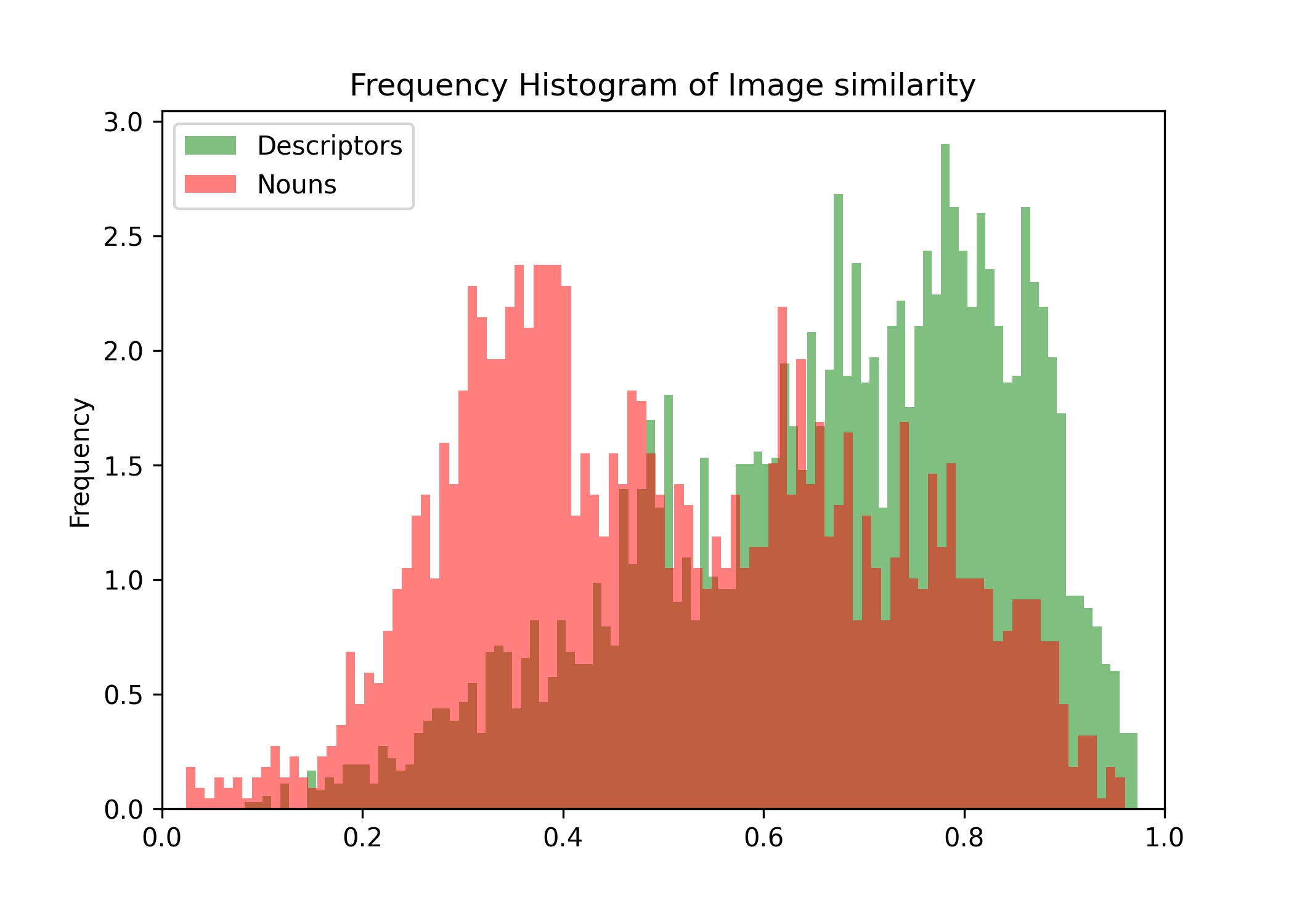}
\captionof{figure}{Descriptor vs Noun Distributions}
\label{fig:Descriptor-vs-Nouns}
\end{minipage}
\end{table}

\subsection{Probing Images and Prompts}

Given a specific random seed for the initial diffusion latents, and a consistent scheduler, images can be produced deterministically by the Stable Diffusion model for a given prompt.
By keeping the same random seed and scheduler but changing the prompt slightly, we are able to produce images which visually show the difference between the 2 prompts. 


To measure similarity between generated images, we experimented with LPIPS (\cite{LPIPS})
perceptual loss (\cite{Johnson2016PerceptualLF}), 
and Watson DFT (\cite{WatsonDFT}) 
measures.


The CLIP model outputs a 768-d embedding tensor each of its 77 input tokens (including padding). 
We measured the semantic difference between prompts by flattening the overall embedding out and using cosine similarity.


In addition, we measured prompt similarity using
Sentence Transformers (\cite{SBERT}) as the text similarity measure,
since these are known to be effective in pure NLP settings. 

%

\section{Experiments \& Observations}

The Stable Diffusion model, supplemented with our chosen metrics, was evaluated 
on over 15,000 image generations.
Over 2,000 prompt variations were used, 
and this enabled us to observe a number of interesting phenomena 
when the prompt styles were isolated, and the base prompt kept constant:

\begin{itemize}
\item Different linguistic categories (such as adjectives, nouns and proper nouns) 
tended to influence the image generation in different, but consistent, ways
\item Simple adjectives (Descriptors) such as descriptions of qualities, 
changes in detail or more ethereal words, 
have a relatively small impact on the generated image 
\item Nouns (on the other hand) tend to dramatically shift the images 
as they introduce new content and act as more than just modifiers
\item The most extreme examples of this can be seen in using 
the names of artists (e.g. ``in the style of Leonardo da Vinci'') which, 
while describing a style,
usually end up dramatically changing the constitution of the image itself 
and therefore making the image much further away from the original 
when measured by image similarity.
\end{itemize}

Indicative average results, and distributions across a wide range of prompts
are illustrated in Table \ref{table:image-similarity} and Figure \ref{fig:Descriptor-vs-Nouns}.
Further experimental results are given in the Appendix.

\section{Discussion}

The technique outlined in this paper allows 
the changes made to text prompts to be quantified in 
conjunction with their effect on the generated images.

We have established that 
words and phrases can be 
categorised, with each category
having a different degree of effect on the overall image.  
While the exact effect of each word or phrase may change from model to model, 
the process for doing this should be robust enough to work on various types of models 
and only require an evaluation to establish baselines for that model going forward.

\section{Future Work}

This work suggests that the similarity of images 
and their CLIP vectors could be used as a form of guidance in image generation in the future, 
potentially being incorporated as part of the loss in the training of these models . 
There is also the intriguing possibility that 
the generative abilities of these kinds of multi-modal models 
could be used to develop and train novel similarity metrics.

\newpage


\section{Ethical Implications}

Generative models such as Stable Diffusion have the potential 
for both good or bad depending on how they are used. 
The models themselves contain various biases towards various groups 
that occur from the images they were trained on and their training procedures. 
This can be seen clearly in experiments on images based on styles from various artists and should be taken into account when evaluating these models and their datasets. 

By documenting these experiments and providing techniques to measure the changes, 
we aim to make it possible to see natural biases in these generative models.

\section{Acknowledgments}

We would like to thank Google and the ML Developer Programs Team for their assistance and compute credits used in the experiments for this paper.

\bibliographystyle{plainnat}
\bibliography{neurips_2022}

\newpage

\appendix

{\LARGE\bf Appendices}

\begin{figure} 
    \centerline{\includegraphics[width=\textwidth]{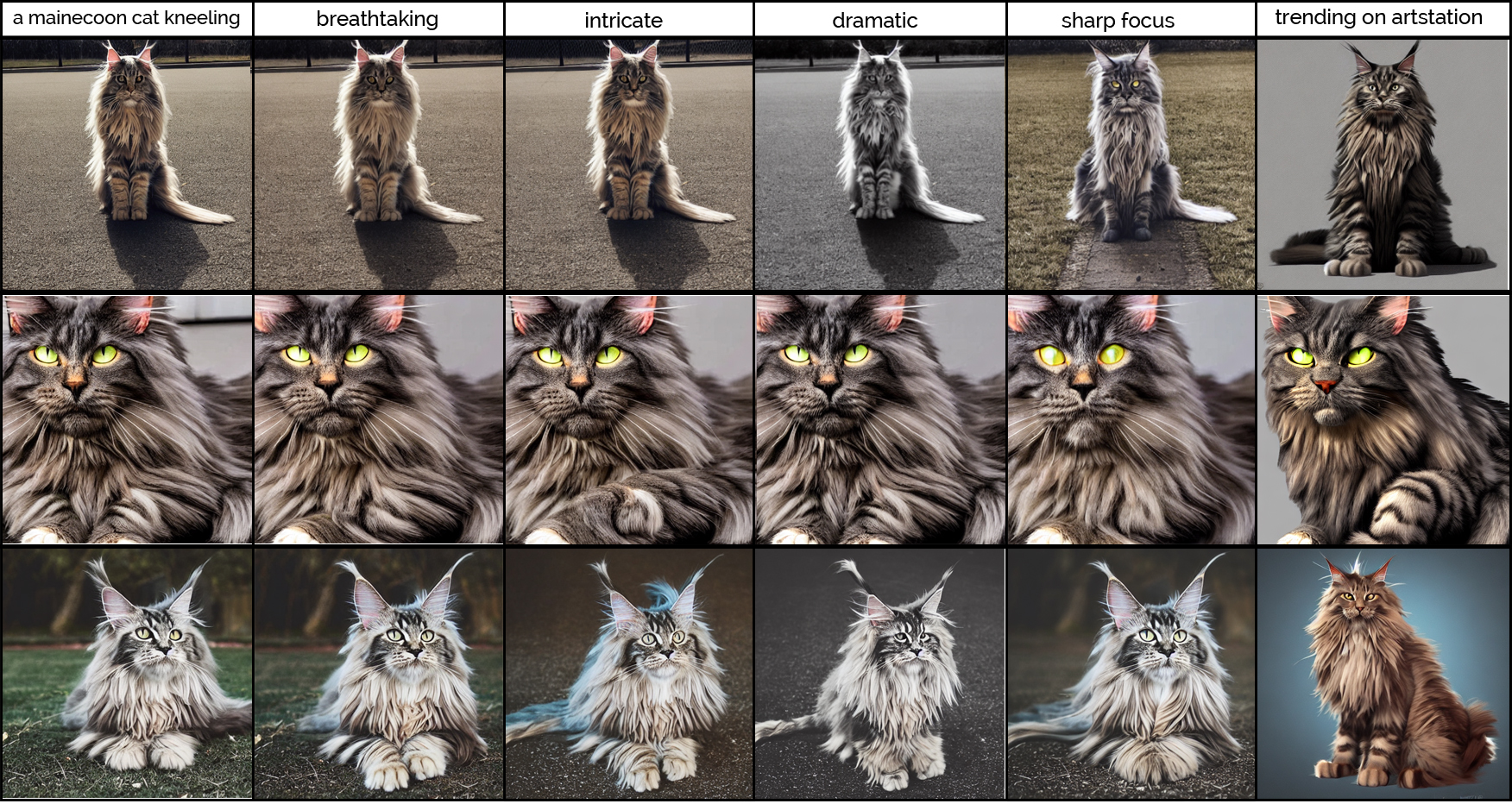}}
    \caption{Overview of different image modifiers (base prompt on left)}
    \label{fig:img512-cg-ts}
\end{figure}

\section{Additional Experiments}

In this section of the Appendix, 
we give illustrative results from our experiments using various types of modifiers 
added to a basic text prompt to measure the impact on the image 
by comparing image similarity to the image made with a base prompt 
(examples are shown in Figure \ref{fig:img512-cg-ts}). 

We also tested additional factors, such as adding 
repeating words,
lighting phrases, 
and 
creating images in the style of a particular artist, 
to see the change they create compared to a base image made with the same seed.

\subsection{Repeating Words}

\begin{figure}[h!]
    \centerline{\includegraphics[width=\textwidth]{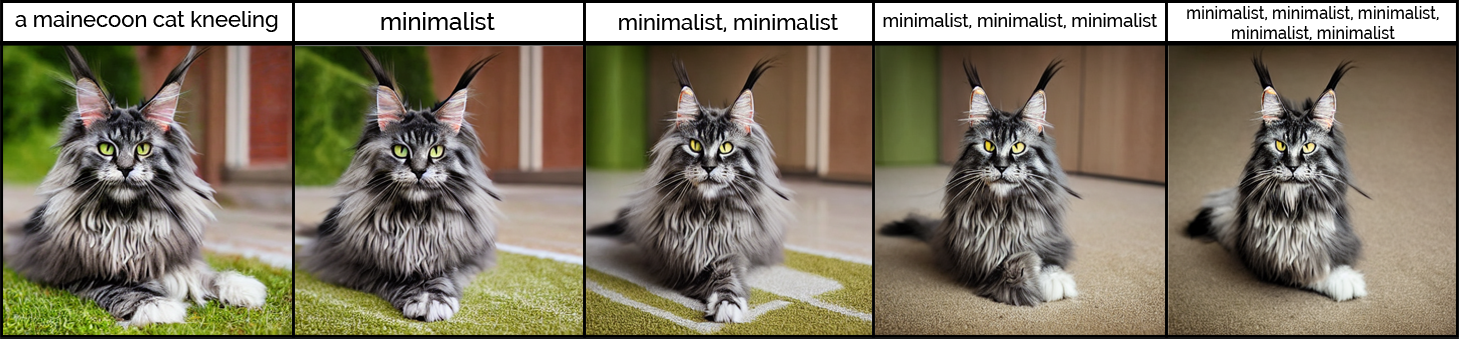}}
    \caption{Repetitions of an image modifier (base prompt on left)}
    \label{fig:repetition}
\end{figure}

One technique that has been reported to improve prompts is that of repeating words.
We analysed repeating modifiers from the descriptor class 
to compare the effects of having the modifier once, 
and also repeated two, three and five times.
Please see Figure \ref{fig:repetition}.
An example of repeating words changing an image can be seen in adding the word ``minimalist''  to the prompt.
Repetition appears to have the effect of removing details from the background 
and eventually (with five occurrences of the word) starts to affect 
the actual subject of the image.
Overall, though, 
we found that while having multiple occurrences of a word can change the image 
it often has no effect at all 
or has an effect that is not a desired semantic effect 
the word might be expected to contribute. 

\newpage

\subsection{Categories of Styles}

As outlined in the main paper, the content of the style section of a prompt can be
viewed within categories.
Figures \ref{fig:individual-descriptors} - \ref{fig:individual-lighting}
show the distribution of similarity \emph{changes} in LPIPS scores for
the different categories of style prompt additions.

As can be seen, adding Descriptors (roughly adjectives or adjectival phrases)
affects the image less than adding Nouns, Artists or Lighting.  
These latter two categories are discussed below.


\begin{table} 
\begin{minipage}[b]{0.48\linewidth}
\centering
    \includegraphics[width=75mm]{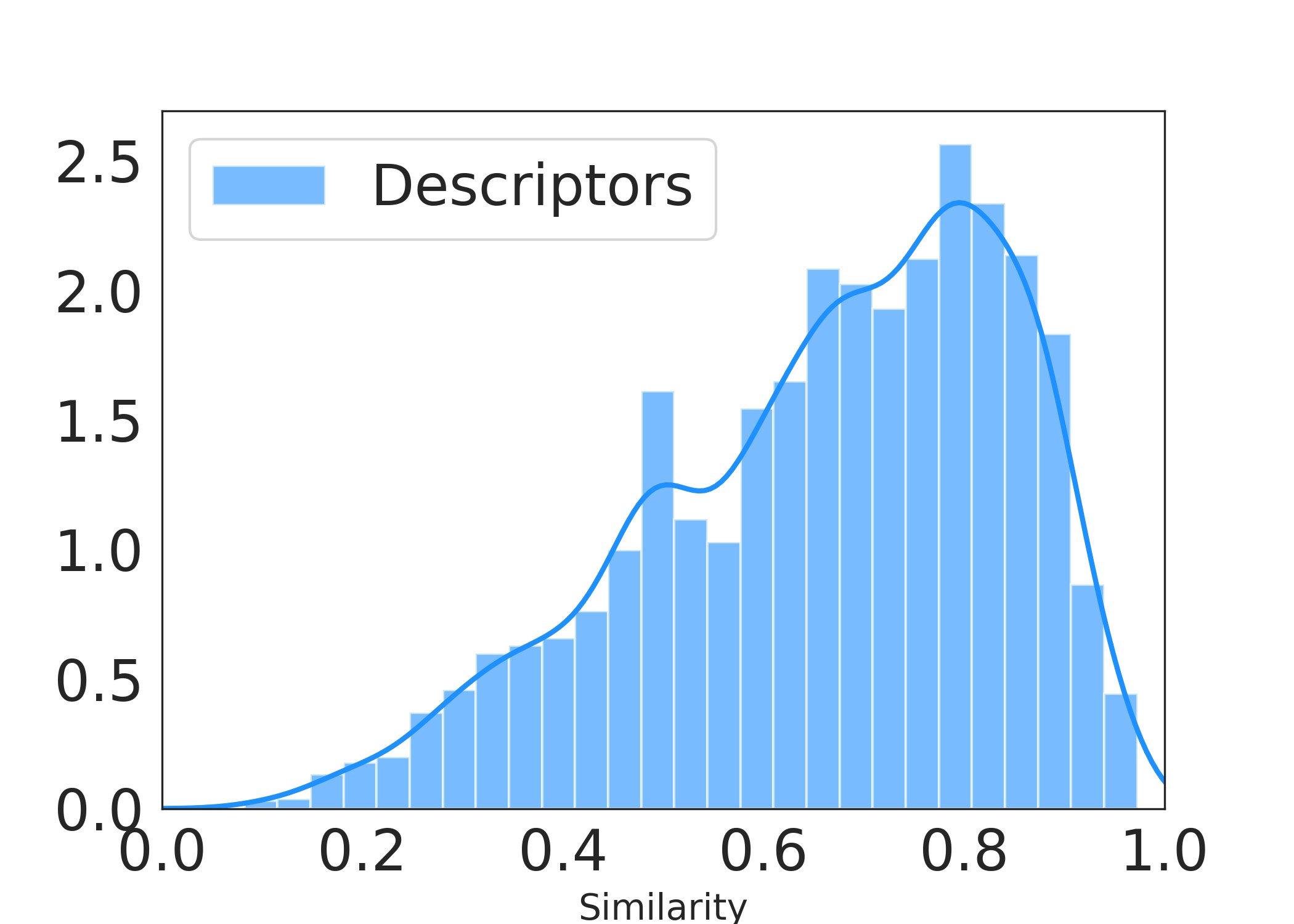} \\[2.0ex]
    \captionof{figure}{Descriptors Distribution}
    \label{fig:individual-descriptors}
\end{minipage}\hfill
\begin{minipage}[b]{0.48\linewidth}
\centering
\includegraphics[width=75mm]{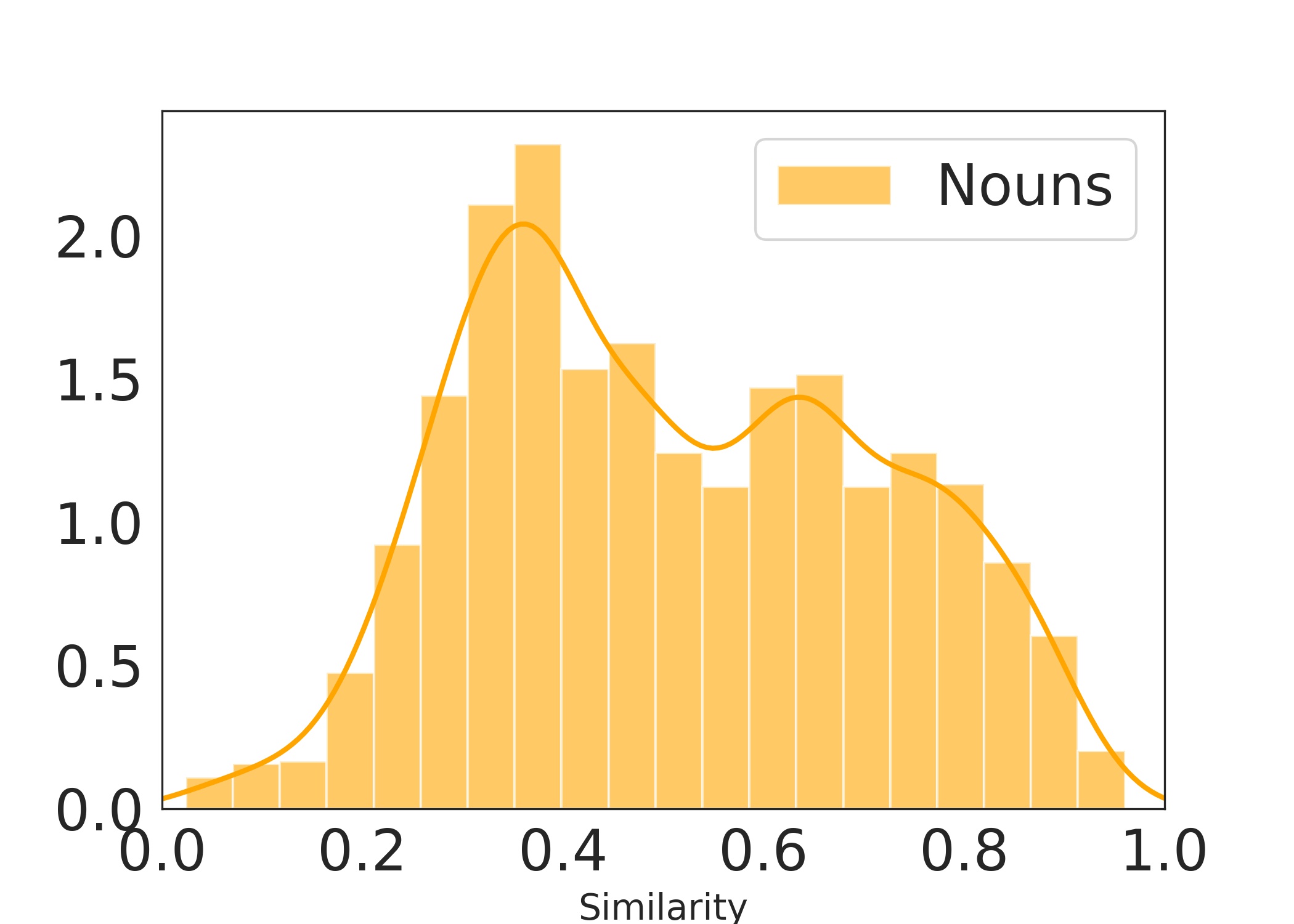} \\[2.0ex]
\captionof{figure}{Nouns Distribution}
\label{fig:individual-nouns}
\end{minipage}
\end{table}

\begin{table} 
\begin{minipage}[b]{0.48\linewidth}
\centering
    \includegraphics[width=75mm]{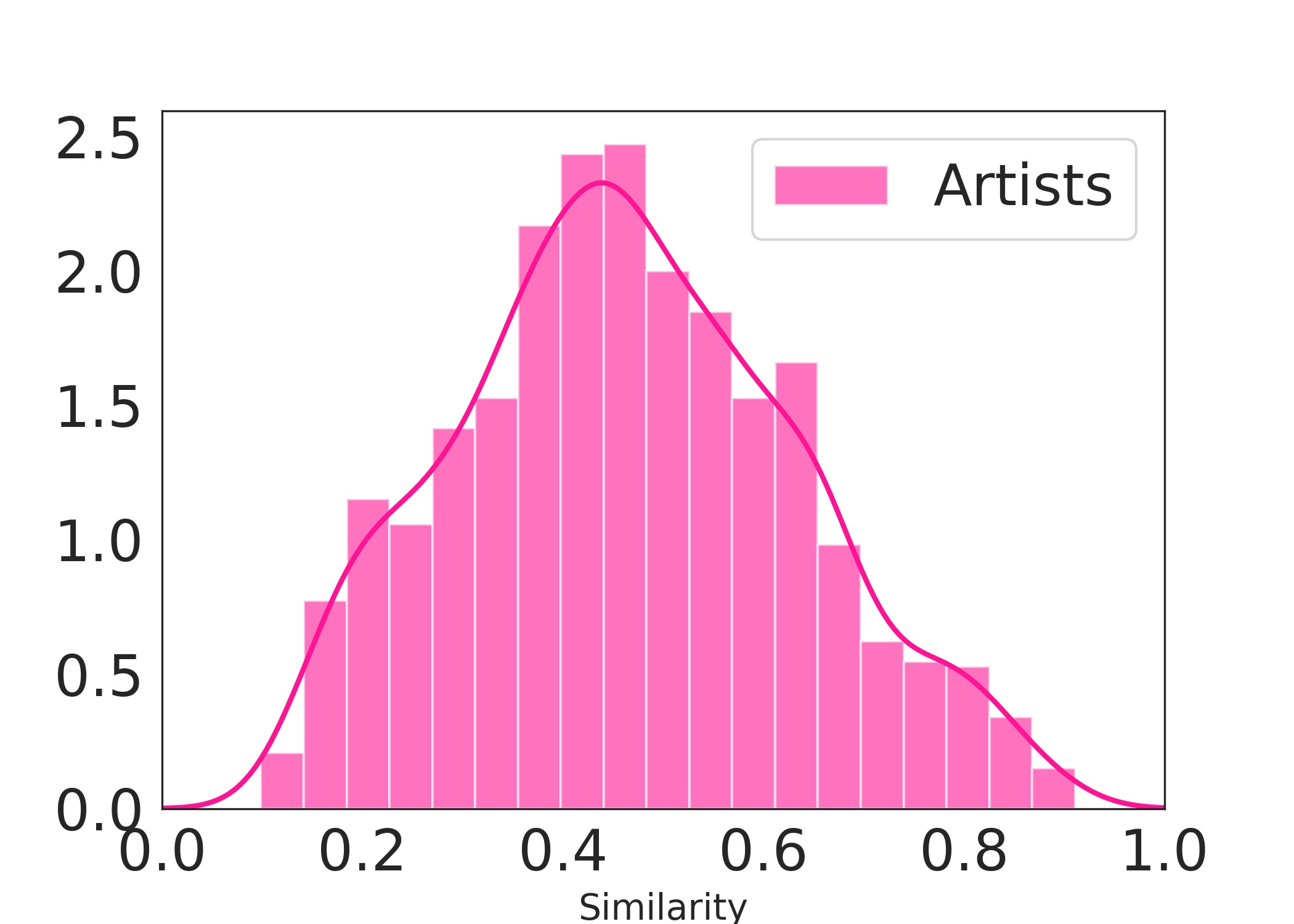} \\[2.0ex]
    \captionof{figure}{Artists Distribution}
    \label{fig:individual-artists}
\end{minipage}\hfill
\begin{minipage}[b]{0.48\linewidth}
\centering
    \includegraphics[width=75mm]{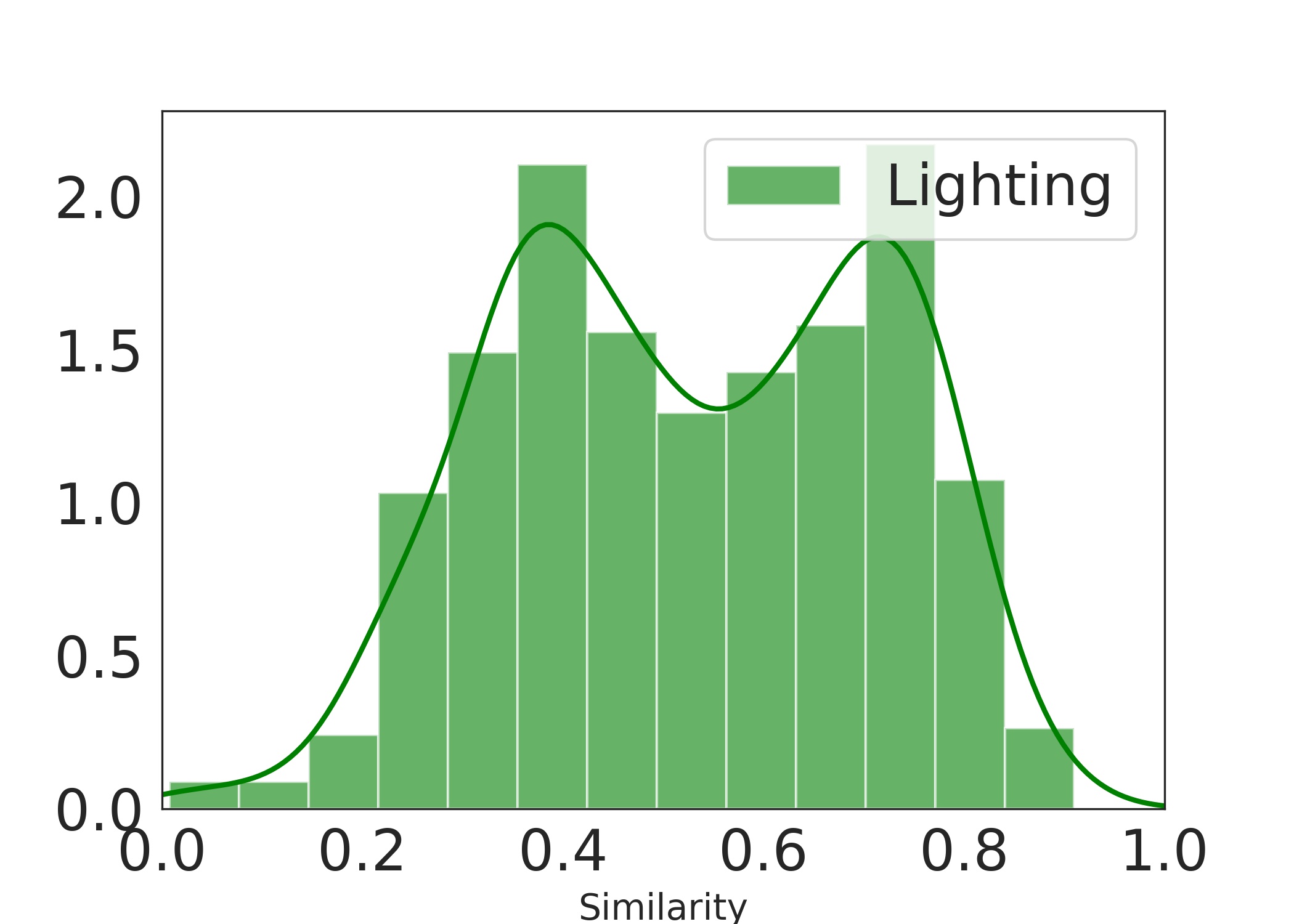} \\[2.0ex]
    \captionof{figure}{Lighting Distribution}
    \label{fig:individual-lighting}
\end{minipage}
\end{table}


\subsection{Adding `Lighting' Words}

\begin{figure}[!hb]
    \centerline{\includegraphics[width=\textwidth]{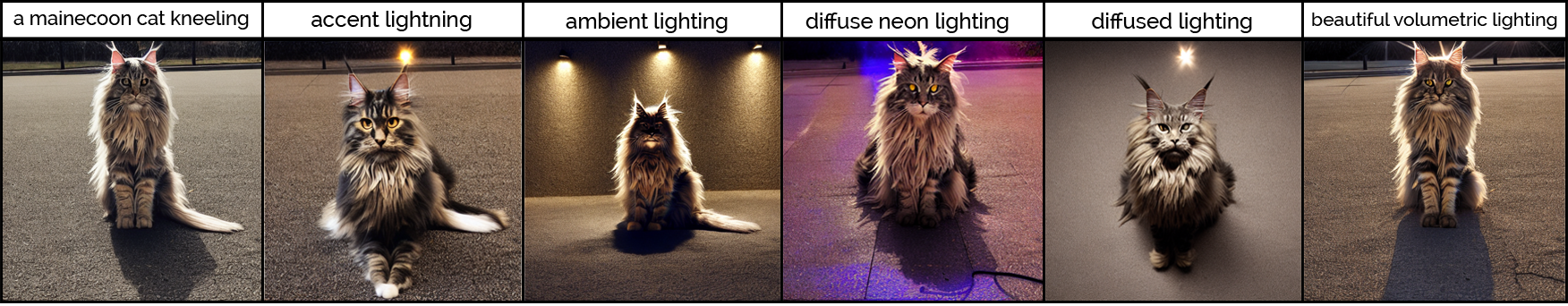}}
    \caption{Comparing lighting modifiers (base prompt on left)}
    \label{fig:lighting-01}
\end{figure}

Words and phrases that describe lighting effects are unusual 
in that they can have either the properties of descriptors 
(which we have found don’t change generated images much)
or they can act as nouns 
(which we observe to make larger changes in the actual content of the image). 

This bimodal behaviour can be seen from the plot of the distribution 
of the image similarities 
in Figure \ref{fig:individual-lighting}, 
which shows 2 clear spikes in the distribution. 
One of these spikes being close to the that of descriptors 
and the other spike being similar to the distribution of nouns.

We also can clearly see this in the images in Figure \ref{fig:lighting-01} 
where we see that phrases such as ``ambient lighting'' 
can change the content significantly, 
whereas a phrase such as ``beautiful volumetric lighting'' 
has relatively little impact on the generated image.
Lighting phrases can often change the look of the subject, 
the mood of the image 
and often just adding lighting information will also change the background of the image.

\begin{figure}[ht]
    \centerline{\includegraphics[width=\textwidth]{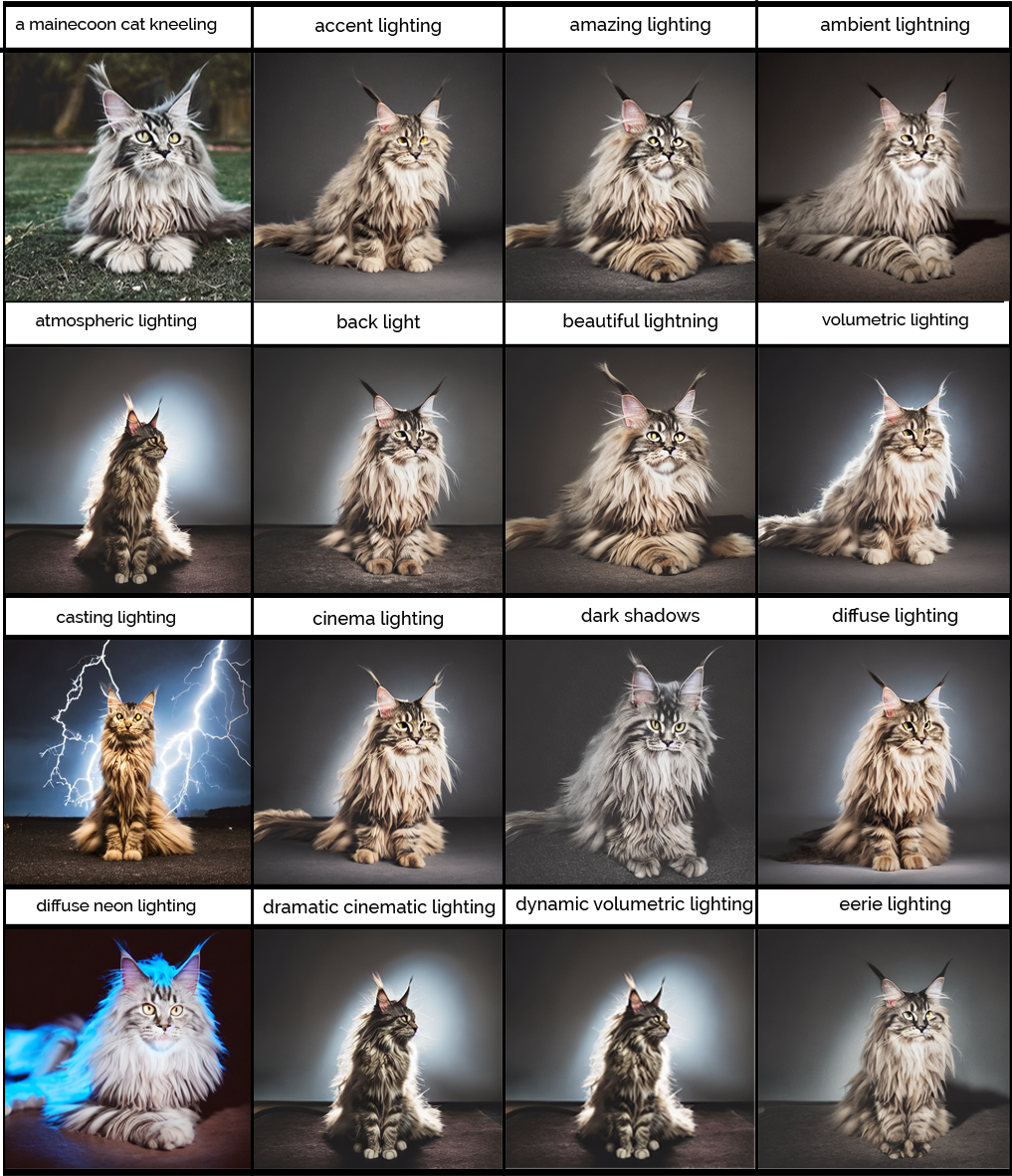}}
    \caption{Examples of lighting prompts added to a single base image}
    \label{fig:lighting-02}
\end{figure}

\newpage

\subsection{Comparing CLIP to LPIPS}

\begin{figure}[!h]
    \centerline{\includegraphics[width=0.9\textwidth]{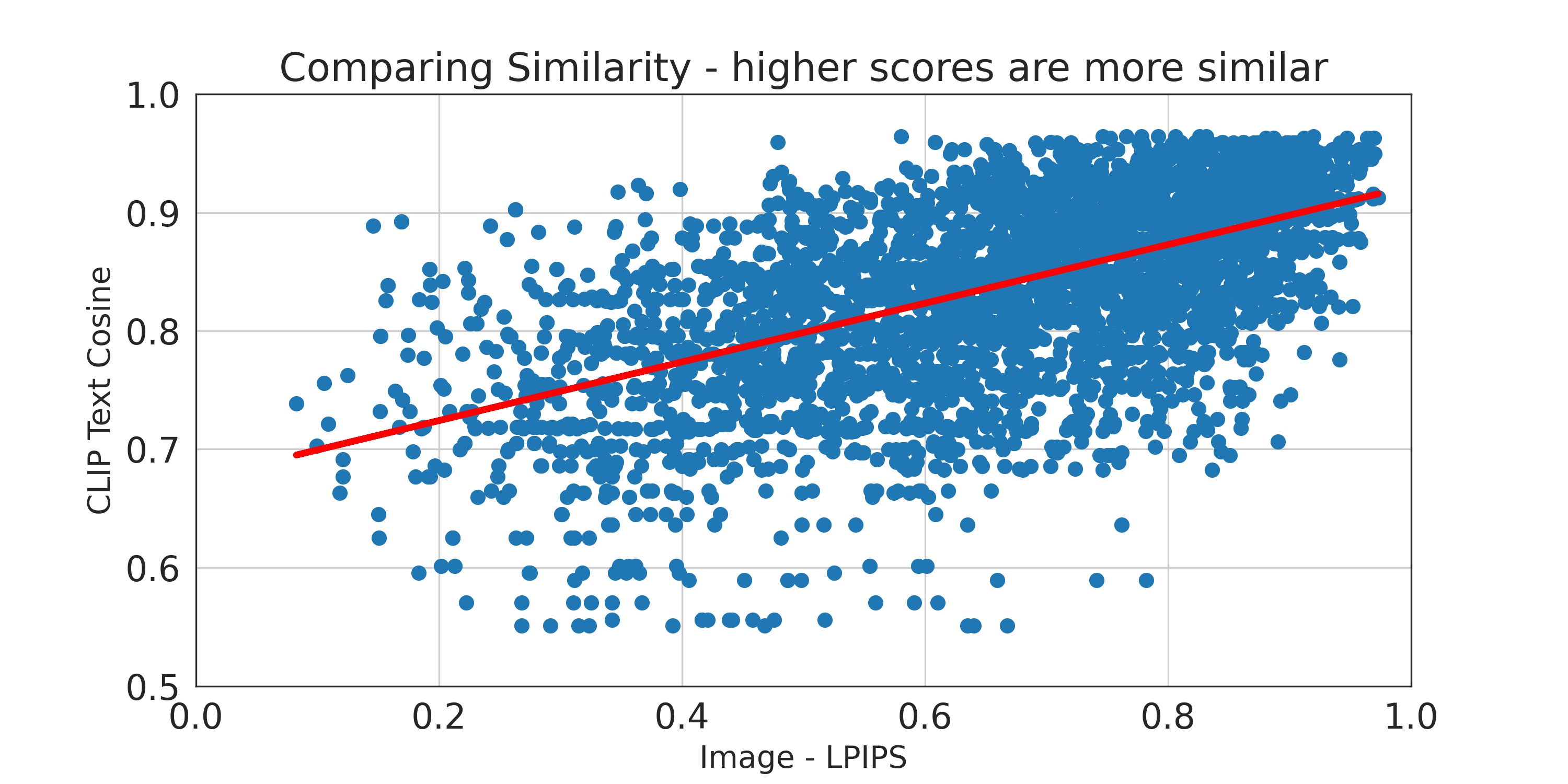}}
    \caption{CLIP Text vs Image score differences for modifiers}
    \label{fig:LPIPS}
\end{figure}

As mentioned in the main paper, 
we examined the relationship between the image similarity measured by comparing 
the images using LPIPS and the text similarity of the prompt.
%
Figure \ref{fig:LPIPS} illustrates how semantic differences in the textual embeddings 
(measured using CLIP cosine distance)
correspond to differences in the images as measured by LPIPS. 

\begin{figure}[!hb]
    \centerline{\includegraphics[width=0.9\textwidth]{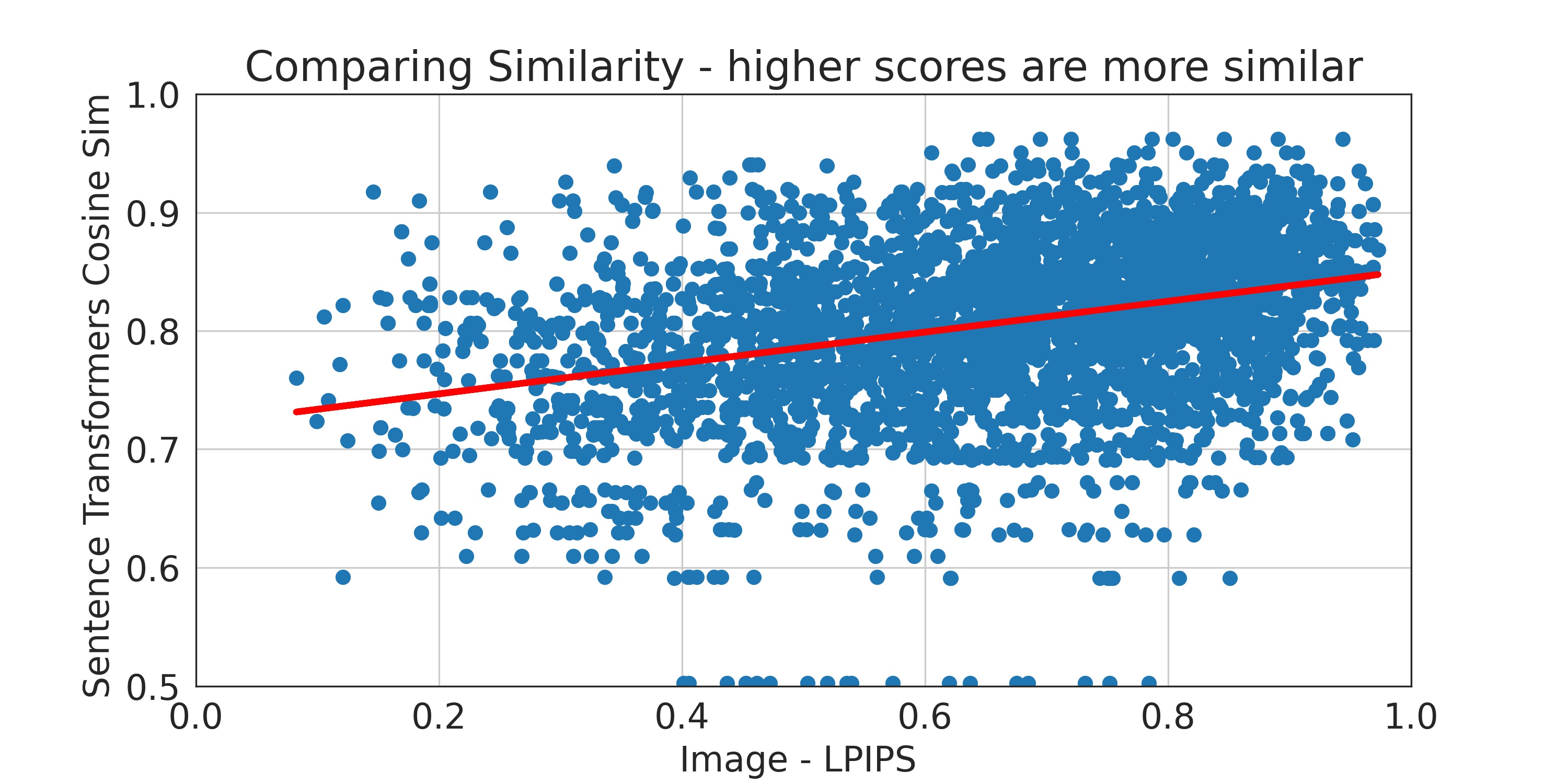}}
    \caption{S-BERT Sentence Embeddings vs Image score differences for modifiers}
    \label{fig:SBert-LPIPS}
\end{figure}

We also performed a comparison of the prompts using Sentence transformers \cite{SBERT}, 
a popular method for text similarity. 
By comparing Figure \ref{fig:SBert-LPIPS} to Figure \ref{fig:LPIPS}, 
it can be seen that the sentence transformer model 
had a relatively weaker correlation to the LPIPS image changes.
This is most likely due to the fact that the CLIP model has been used 
in the training of the overall model and so its representations 
are naturally aligned to match the generation of images.


\newpage

\subsection{Styled by Artist}

We examined the effect of adding the prompt ``in the style of'' 
along with an artist's name to the original prompt with an extensive set of
artist names.

Qualitatively, we found that including an artist’s name in the prompt 
can lead to image generation changes 
on multiple levels,
with the artist's name alone potentially informing the image generation as to 
the art medium (photo, oil paint, water color etc), 
the color palette 
and often the racial qualities of the subject.  

\begin{figure}[!ht]
\centerline{\includegraphics[width=\textwidth]{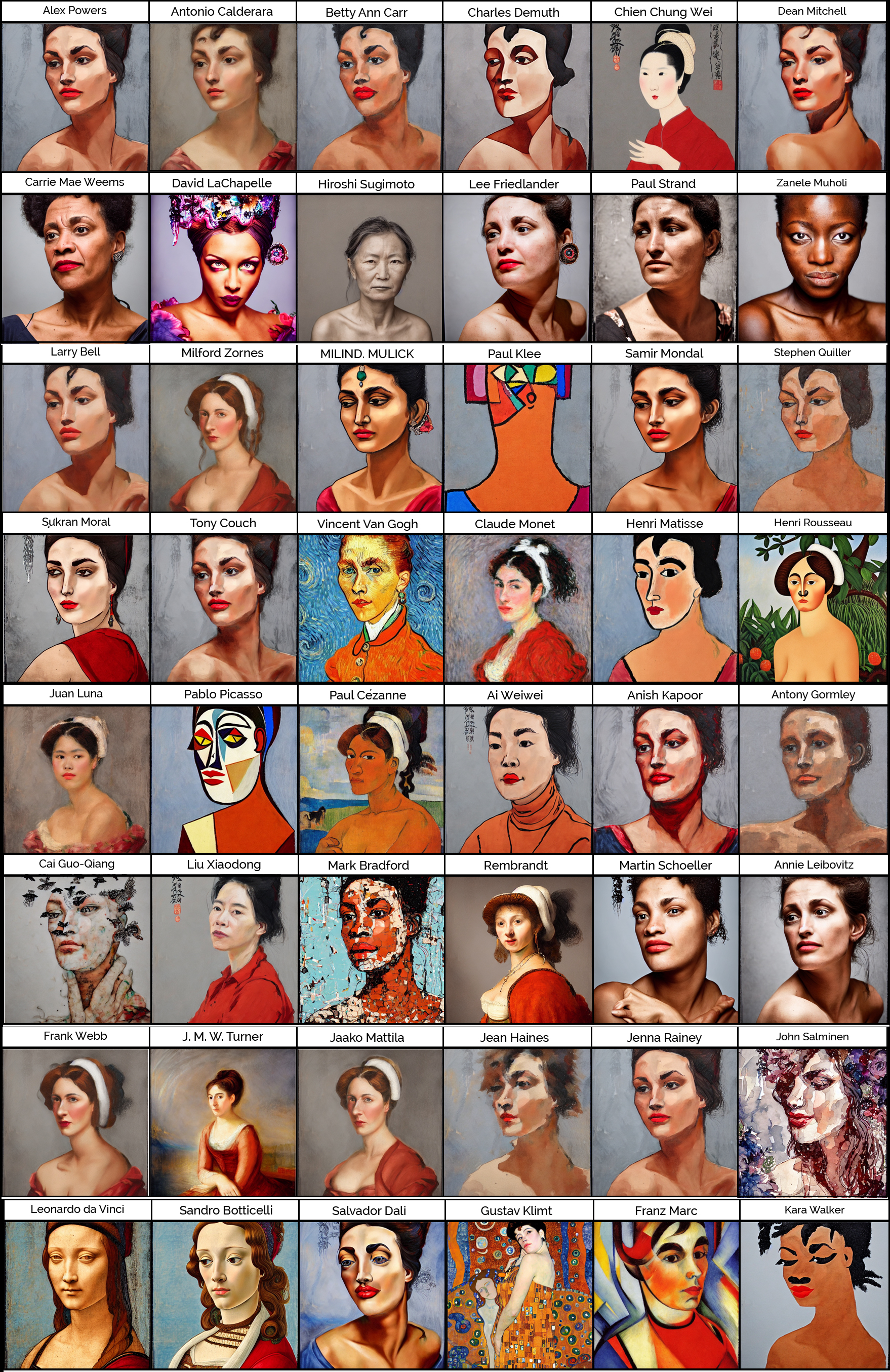}}
\caption{``A portrait of a beautiful woman in the style of ARTIST NAME'' 
}
\label{fig:artists-main}
\end{figure}

Please see Figure \ref{fig:artists-main} for an illustration of the range of
effects seen.

\newpage

\newpage

\section{Guidance for choosing your Prompts}

While creating a prompt to generate an image may initially seem trivial and easy, 
it very quickly becomes clear to aspiring artists that it is often much harder 
that it might first appear
to create a prompt that produces an output that matches their original intent.

Here we propose a procedure for creating prompts, 
where each stage builds on the previous ones
:

\begin{itemize}
    \item Start with a clear noun-based statement that contains the main subject you wish to have in the image.
    \item Be systematic about recording what seeds are effective, 
    since it is easier to iterate on images based on a firm foundation 
    (as can be seen from the samples included in this Appendix).
    \item The properties of the Stable Diffusion model allow it to be very good 
    at replicating style of various artists. 
    Because of this, it is often beneficial to look for artists 
    and key styles that you would like to emulate and add that to your prompt.
    \item Finally, experiment with descriptors that you wish to add, 
    considering the importance of (for instance) having 
    a simple or a busy/intricate background. 
    This can be done through adding lighting effects phrases and repeating words.
\end{itemize}

Overall, the process for `Prompt Engineering' can also be seen as an artistic endeavour,
and the techniques required for success must be honed by over time, 
using experimentation, imagination and creativity.

\end{document}